\documentclass[runningheads, orivec]{llncs}
\usepackage[T1]{fontenc}
\usepackage[utf8]{inputenc}
\usepackage[T1]{fontenc}

\usepackage{amsmath}
\usepackage{amssymb}
\usepackage{bm}          

\usepackage{graphicx}
\usepackage{subfig}      

\usepackage{booktabs}    
\usepackage{multirow}

\usepackage{algorithm}
\usepackage{algorithmic} 

\usepackage{hyperref}
\usepackage{cite}        

\usepackage{xcolor}      
\usepackage{microtype}   

\usepackage{booktabs}
\usepackage{tabularx}

\usepackage{color}

\begin{document}

\title{Uncertainty Estimation for Deep Reconstruction in Actuatic Disaster Scenarios with Autonomous Vehicles}

\titlerunning{Unc. Estimation for Deep Recon. in Actuatic Disasters}
%
\author{Samuel Yanes Luis\inst{1} \and
    Alejandro Casado Pérez\inst{2} \and
    Alejandro Mendoza Barrionuevo\inst{2} \and
    Dame Seck Diop\inst{2} \and
    Sergio Toral Marín\inst{2} \and
    Daniel Gutiérrez Reina\inst{2}
}
\authorrunning{S. Yanes et al.}
%
\institute{Department of Electronic Technology, University of Sevilla, Av. de la Reina Mercedes S/N, 41012, Sevilla, Spain. \and
    \email{syanes@us.es}\\
    \and
    Department of Electronic Engineering, University of Sevilla, Av. de los Descubrimientos S/N, 41005, Sevilla, Spain. \\
    \email{\{acasado4, amendoza1, dseck, dgutierrezreina, storal\}@us.es}}
\maketitle              

\begin{abstract}
Accurate reconstruction of environmental scalar fields from sparse 
onboard observations is essential for autonomous vehicles engaged in 
aquatic monitoring. Beyond point estimates, principled uncertainty 
quantification is critical for active sensing strategies such as 
Informative Path Planning, where epistemic uncertainty drives data 
collection decisions. This paper compares Gaussian Processes, Monte 
Carlo Dropout, Deep Ensembles, and Evidential Deep Learning for 
simultaneous scalar field reconstruction and uncertainty 
decomposition under three perceptual models representative of 
real sensor modalities. Results show that Evidential Deep Learning 
achieves the best reconstruction accuracy and uncertainty calibration 
across all sensor configurations at the lowest inference cost, while 
Gaussian Processes are fundamentally limited by their stationary 
kernel assumption and become intractable as observation density grows. 
These findings support Evidential Deep Learning as the preferred 
method for uncertainty-aware field reconstruction in real-time 
autonomous vehicle deployments.

\end{abstract}

\section{Introduction}

Deploying autonomous vehicles for environmental monitoring is, 
at its core, a decision-making problem under uncertainty. A 
surface vessel or aerial drone tasked with mapping a marine 
hydrocarbon spill must continuously answer two intertwined 
questions: \emph{what does the contaminant field look like 
right now?} and \emph{where is that estimate least trustworthy?} 
The first question drives reconstruction; the second drives 
action. Without a principled answer to both, any adaptive 
monitoring strategy is blind~\cite{mansfield2024survey}.

This coupling between estimation and uncertainty is especially 
acute for phenomena that are chaotic in nature. Oil-spill 
dynamics, algal blooms, or chemical plumes can be simulated 
via advection-diffusion equations, yet their real-world 
realisation is sensitive to unobserved boundary conditions, 
such that reconstruction algorithms must express calibrated 
ignorance over unvisited regions and distinguish between 
\emph{epistemic uncertainty} arising from lack of data and 
\emph{aleatoric uncertainty} from irreducible sensor 
noise~\cite{amini2020deep}. Only the former is reducible by 
collecting additional measurements and therefore actionable 
for path planning. This challenge is amplified by platform 
heterogeneity: autonomous surface vehicles with contact 
sensors deliver scalar readings at isolated points, while 
aerial vehicles with multispectral or thermal cameras return 
spatially dense but indirect observations over wide 
areas~\cite{zakaria2021uav}. Each observation model induces 
a different geometry of information and uncertainty structure, 
and a practical framework must accommodate all of them.

Gaussian processes have long been the standard tool for this 
setting precisely because uncertainty is native to their 
formulation: posterior variance is available in closed form 
and well-calibrated by construction~\cite{mansfield2024survey}. 
Their limitations are equally well-known: cubic scaling with 
the number of observations and an inability to exploit the 
rich spatial structure encoded in image-like measurements. 
Deep learning addresses both shortcomings, as networks trained 
on physics-based synthetic data generalise across observation 
modalities and run in milliseconds at inference time, but 
standard deterministic architectures produce point estimates 
with no notion of confidence~\cite{laksh2017deep}. Recovering 
calibrated uncertainty from deep models has therefore become 
a central research question. Monte Carlo 
Dropout~\cite{gal2016dropout} approximates a Bayesian 
posterior by averaging stochastic forward passes; Deep 
Ensembles~\cite{laksh2017deep} derive uncertainty from the 
disagreement of independently trained networks; and Evidential 
Deep Learning~\cite{amini2020deep} places a higher-order 
evidential prior over the likelihood, providing both 
uncertainty types in a single forward pass without sampling.

To the best of our knowledge, this is the first systematic comparison of these four families under heterogeneous observation models 
in an environmental-monitoring setting. This gap matters is important 
because the utility of uncertainty estimates in active monitoring depends 
not only on their average calibration, but on whether 
epistemic uncertainty specifically correlates with prediction 
error, so that it can serve as a reliable criterion for 
directing the vehicle towards maximally informative 
locations~\cite{mansfield2024survey,chen2019robotic}. We present a comparative study of Gaussian Process regression, Monte Carlo 
Dropout, Deep Ensembles, and Evidential Deep Learning for 
simultaneous field reconstruction and uncertainty estimation 
in autonomous environmental monitoring. Concretely: 
\emph{(i)}~we formulate the problem under a unified 
observation-model framework covering real-world sensor 
modalities typical of aerial and surface platforms; 
\emph{(ii)}~we evaluate reconstruction accuracy and 
uncertainty calibration, with explicit separation of 
aleatoric and epistemic components, on physics-based 
oil-spill simulations; and \emph{(iii)}~we analyse the 
correlation between epistemic uncertainty and prediction 
error as an indicator of active-learning suitability. 

The paper is organised as follows. Section~\ref{sec:related} 
reviews related work. Section~\ref{sec:problem} formalises 
the problem. Section~\ref{sec:methods} describes estimation 
and observation models. Section~\ref{sec:results} presents 
results, and Section~\ref{sec:conclusion} concludes.

\section{Related Work}
\label{sec:related}

Gaussian Processes have been widely adopted for scalar field 
estimation in autonomous vehicle missions due to their principled 
probabilistic formulation~\cite{samaniego2021bayesian}. In the context of aquatic 
monitoring, GPs provide a natural mechanism for uncertainty 
quantification: the posterior variance directly reflects the 
confidence of the model at unobserved locations, making them a 
natural fit for IPP strategies that seek to reduce uncertainty 
along the planned path~\cite{syanes2023censored}. However, classical GP formulations 
rely on a fixed, stationary kernel, typically the Radial Basis 
Function (RBF) , which imposes isotropy and stationarity on the 
estimated field. This implies a homoscedastic noise assumption, where 
the aleatoric uncertainty is treated as a global hyperparameter 
fitted by maximizing the marginal log-likelihood, rather than being 
learned as a spatially varying quantity. As a result, GPs struggle 
to capture complex, non-stationary phenomena or sensor noise 
patterns that vary across the environment~\cite{syanes2024deep}.

Deep generative models, particularly Variational 
Autoencoders, have been explored for reconstructing 
environmental phenomena such as oil spills and floating 
debris~\cite{vae}. Trained on physics-based simulators, 
these models capture complex non-stationary spatial 
patterns that GPs struggle to represent. However, 
predictive uncertainty is not a native output of these 
architectures, which restricts their use in active 
sensing frameworks. This motivates equipping similarly 
flexible deep architectures with principled uncertainty 
estimation, as pursued in this work.

Several methods have been proposed to equip deep neural networks 
with uncertainty estimation capabilities. Monte Carlo 
Dropout~\cite{gal2016dropout} interprets dropout at inference time as approximate 
Bayesian inference, yielding a simple yet effective way to obtain 
epistemic uncertainty estimates. Deep Ensembles~\cite{laksh2017deep} train 
multiple independent networks and exploit their predictive 
disagreement as a proxy for uncertainty, achieving strong empirical 
calibration at the cost of increased computation. Evidential Deep 
Learning~\cite{amini2020deep} takes a different approach by placing a conjugate 
prior over the output distribution, enabling closed-form decomposition 
of epistemic and aleatoric uncertainty from a single forward pass.

Despite their demonstrated effectiveness, these methods have been 
evaluated primarily in the context of image classification tasks, 
with comparatively little attention paid to regression problems. 
More importantly, their behavior under the perceptual models 
characteristic of autonomous vehicles , including cone-shaped 
fields of view, point-wise sensors, and nadir-looking cameras, 
has not been systematically studied. The feasibility and calibration 
of these methods under such observation models, and their suitability 
for integration into IPP pipelines for USVs and UAVs, remains an 
open question that this work aims to address.

\section{Problem Statement}
\label{sec:problem}

\subsection{Scalar Field Reconstruction}

Let $\mathcal{X} \subset \mathbb{R}^2$ denote the spatial domain of
interest, and let $f: \mathcal{X} \rightarrow \mathbb{R}$ be an
unknown scalar field representing the environmental phenomenon to
be monitored (e.g., pollutant concentration or debris density).
We assume that $f$ is static or quasi-static over the duration of
the mission.

A vehicle collects a set of $N$ observations
$\mathcal{D} = \{(\mathbf{x}_i, y_i)\}_{i=1}^{N}$, where
$\mathbf{x}_i \in \mathcal{X}$ denotes the measurement location and
$y_i \in \mathbb{R}$ is the corresponding observation. The observation
is related to the true field through a perceptual model
$g: \mathcal{X} \rightarrow \mathcal{Y}$, which encodes the
sensor's measurement process:
\begin{equation}
    y_i = g(f, \mathbf{x}_i) + \varepsilon_i,
    \label{eq:obs_model}
\end{equation}
where $\varepsilon_i$ represents sensor noise. The perceptual model
$g$ depends on the sensor modality: a point-wise measurement
(e.g., a contact water quality sensor), a cone-shaped field of view
(e.g., a lateral camera for debris detection), or a nadir-looking
footprint (e.g., an aerial camera). In all cases, $g$ defines how
the true field is integrated or aggregated into a scalar observation
at each vehicle pose.

Rather than seeking a point estimate $\hat{f}$, the primary
objective is to learn a predictive distribution
$p(f(\mathbf{x}) \mid \mathbf{x}, \mathcal{D})$ for each location
$\mathbf{x} \in \mathcal{X}$. We model this distribution as a
Gaussian,
\begin{equation}
    p\bigl(f(\mathbf{x}) \mid \mathbf{x}, \mathcal{D}\bigr)
    = \mathcal{N}\!\left(\hat{f}(\mathbf{x}),\,
    \hat{\sigma}^2(\mathbf{x})\right),
    \label{eq:gaussian_assumption}
\end{equation}
where the mean $\hat{f}(\mathbf{x})$ serves as the point
reconstruction and $\hat{\sigma}^2(\mathbf{x})$ captures the total
predictive uncertainty. This Gaussian assumption is standard in
heteroscedastic regression~\cite{nix1994estimating} and is well-suited
to the present setting: the sensor noise $\varepsilon_i$ in
Eq.~\ref{eq:obs_model} is physically reasonable to model as Gaussian,
and the epistemic component reflects parameter uncertainty that, under
ensemble averaging, also converges to a Gaussian marginal.

Under this assumption, the model is trained by minimising the
Negative Log-Likelihood~(NLL) over the collected observations:
\begin{equation}
    \mathcal{L}_{\text{NLL}} = \frac{1}{N}\sum_{i=1}^{N}
    \left[
        \frac{\bigl(y_i - \hat{f}(\mathbf{x}_i)\bigr)^2}
             {2\,\hat{\sigma}^2(\mathbf{x}_i)}
        + \frac{1}{2}\log \hat{\sigma}^2(\mathbf{x}_i)
    \right].
    \label{eq:nll_train}
\end{equation}
This objective jointly incentivises accurate reconstruction and
well-calibrated predictive variance: overestimating
$\hat{\sigma}^2$ is penalised by the log term, while underestimating
it inflates the squared residual. This contrasts with
reconstruction-only approaches trained with mean squared
error~(MSE), where any inferred variance is disconnected from the
training signal and collapses to near-zero values, rendering
uncertainty estimates unreliable~\cite{vae}.

\subsection{Uncertainty Decomposition}

Beyond the total predictive variance, we seek to characterise the
separate sources of uncertainty associated with $\hat{f}$.
Following the standard decomposition~\cite{amini2020deep}, the total
predictive uncertainty at a location $\mathbf{x}$ can be written as:
\begin{equation}
    \sigma^2_{\text{total}}(\mathbf{x}) =
    \underbrace{\sigma^2_{\text{ale}}(\mathbf{x})}_{\text{aleatoric}} +
    \underbrace{\sigma^2_{\text{epi}}(\mathbf{x})}_{\text{epistemic}},
    \label{eq:uncertainty}
\end{equation}
where aleatoric uncertainty $\sigma^2_{\text{ale}}$ reflects
irreducible noise inherent to the measurement process, and epistemic
uncertainty $\sigma^2_{\text{epi}}$ reflects the model's lack of
knowledge due to limited observations, which can in principle be
reduced by collecting additional data. How $\hat{\sigma}^2(\mathbf{x})$
is decomposed into these two components depends on the estimation
method and is detailed in Section~\ref{sec:methods}.

This decomposition is central to the active sensing use case: only
epistemic uncertainty is actionable for path planning, as aleatoric
uncertainty remains invariant to further sampling. A method that
conflates both sources provides a misleading signal for guiding data
collection.

\section{Methods}
\label{sec:methods}

\subsection{Reconstruction Backbone: U-Net Architecture}

All deep learning methods share a common U-Net 
backbone~\cite{UNET} that takes as input two spatial matrices 
of size $H \times W$: an observation mask 
$\mathbf{M} \in \{0,1\}^{H \times W}$ and a value matrix 
$\mathbf{V} \in \mathbb{R}^{H \times W}$, concatenated along 
the channel dimension to form 
$\mathbf{X} = [\mathbf{M}, \mathbf{V}] \in 
\mathbb{R}^{2 \times H \times W}$. The encoder comprises 4 
blocks of two convolutional layers with batch normalization 
and ReLU activations followed by max-pooling. The decoder mirrors this structure 
via transposed convolutions with skip connections, and a final 
$1 \times 1$ convolution produces the reconstructed field 
$\hat{\mathbf{F}} \in \mathbb{R}^{H \times W}$.

\subsection{Monte Carlo Dropout}

Monte Carlo Dropout~\cite{gal2016dropout} inserts dropout 
layers with rate $p$ after each convolutional block. 
At inference, dropout remains active and $T$ stochastic 
forward passes yield predictions 
$\{\hat{\mathbf{F}}^{(t)}\}_{t=1}^{T}$, from which epistemic 
uncertainty and the predictive mean are estimated as:
\begin{equation}
    \hat{f}(\mathbf{x}) = \frac{1}{T}\sum_{t=1}^{T} 
    \hat{f}^{(t)}(\mathbf{x}), \qquad
    \sigma^2_{\text{epi}}(\mathbf{x}) = \frac{1}{T}
    \sum_{t=1}^{T}\left(\hat{f}^{(t)}(\mathbf{x}) - 
    \hat{f}(\mathbf{x})\right)^2.
    \label{eq:mcd}
\end{equation}
Aleatoric uncertainty is obtained from an auxiliary 
log-variance head trained with the heteroscedastic NLL loss:
\begin{equation}
    \mathcal{L}_{\text{NLL}} = \frac{1}{|\mathcal{X}|}
    \sum_{\mathbf{x}} \left[ 
    \frac{\left(f(\mathbf{x}) - \hat{f}(\mathbf{x})\right)^2}
    {2\hat{\sigma}^2_{\text{ale}}(\mathbf{x})} + 
    \frac{1}{2}\log \hat{\sigma}^2_{\text{ale}}(\mathbf{x}) 
    \right],
    \label{eq:mcd_loss}
\end{equation}

\subsection{Deep Ensembles}

Deep Ensembles~\cite{laksh2017deep} train $M$ independent 
backbone instances from different random seeds, each with 
an auxiliary log-variance head optimized with 
Eq.~\eqref{eq:mcd_loss}. At inference, epistemic uncertainty 
is derived from inter-member disagreement and aleatoric 
uncertainty is averaged across members:
\begin{equation}
    \hat{f}(\mathbf{x}) = \frac{1}{M}\sum_{m=1}^{M} 
    \hat{f}^{(m)}(\mathbf{x})
    \label{eq:ens}
\end{equation}

\begin{equation}
    \sigma^2_{\text{epi}}(\mathbf{x}) = \frac{1}{M}
    \sum_{m=1}^{M}\left(\hat{f}^{(m)}(\mathbf{x}) - 
    \hat{f}(\mathbf{x})\right)^2, \quad
    \hat{\sigma}^2_{\text{ale}}(\mathbf{x}) = 
    \frac{1}{M}\sum_{m=1}^{M} 
    \hat{\sigma}^{2(m)}_{\text{ale}}(\mathbf{x})
    \label{eq:ens2}
\end{equation}

\subsection{Evidential Deep Learning}

Evidential Deep Learning~\cite{amini2020deep} places a 
Normal-Inverse-Gamma prior over the likelihood, modifying 
the backbone to output four quantities 
$(\gamma, \nu, \alpha, \beta)$ per spatial location, 
with constraints $\nu > 0$, $\alpha > 1$, $\beta > 0$. 
Uncertainty estimates are available in closed form:
\begin{equation}
    \hat{f}(\mathbf{x}) = \gamma, \quad
    \sigma^2_{\text{epi}}(\mathbf{x}) = 
    \frac{\beta}{\nu(\alpha-1)}, \quad
    \sigma^2_{\text{ale}}(\mathbf{x}) = 
    \frac{\beta}{\alpha-1},
    \label{eq:edl}
\end{equation}
where we omit the explicit dependence on $\mathbf{x}$ for 
brevity. Training minimizes the NIG log-likelihood plus an 
evidence regularizer that penalizes high evidence on 
incorrect predictions:
\begin{equation}
    \mathcal{L}_{\text{EDL}} = \sum_{\mathbf{x}} \left[
    \mathcal{L}_{\text{NIG}}(\mathbf{x}) + 
    \lambda \cdot |f(\mathbf{x}) - \gamma(\mathbf{x})| 
    \cdot \left(2\nu(\mathbf{x}) + \alpha(\mathbf{x})\right)
    \right],
    \label{eq:edl_loss}
\end{equation}
with $\lambda = 10^{-3}$ and gradient clipping at max norm 1.0.

\subsection{Gaussian Process Baseline}

As a probabilistic baseline, we employ a Gaussian 
Process~\cite{rasmussen2006gaussian} with composite kernel:
\begin{equation}
    k(\mathbf{x}, \mathbf{x}') = \sigma_f^2 \exp\!\left(
    -\frac{\|\mathbf{x} - \mathbf{x}'\|^2}{2\ell^2}\right)
    + \sigma_n^2 \,\delta(\mathbf{x}, \mathbf{x}'),
    \label{eq:rbf_white}
\end{equation}
with hyperparameters $\{\sigma_f^2, \ell, \sigma_n^2\}$ 
optimized by maximizing the marginal log-likelihood per 
test sample. The posterior variance captures epistemic 
uncertainty through contraction at observed locations, 
while $\sigma_n^2$ provides a spatially uniform aleatoric 
estimate — a homoscedastic limitation that, as shown in 
Section~\ref{sec:results}, proves critical for fields 
with heterogeneous spatial regularity.

\subsection{Perceptual Models}

\begin{figure}[htbp]
    \centering
    \includegraphics[width=0.9\textwidth]{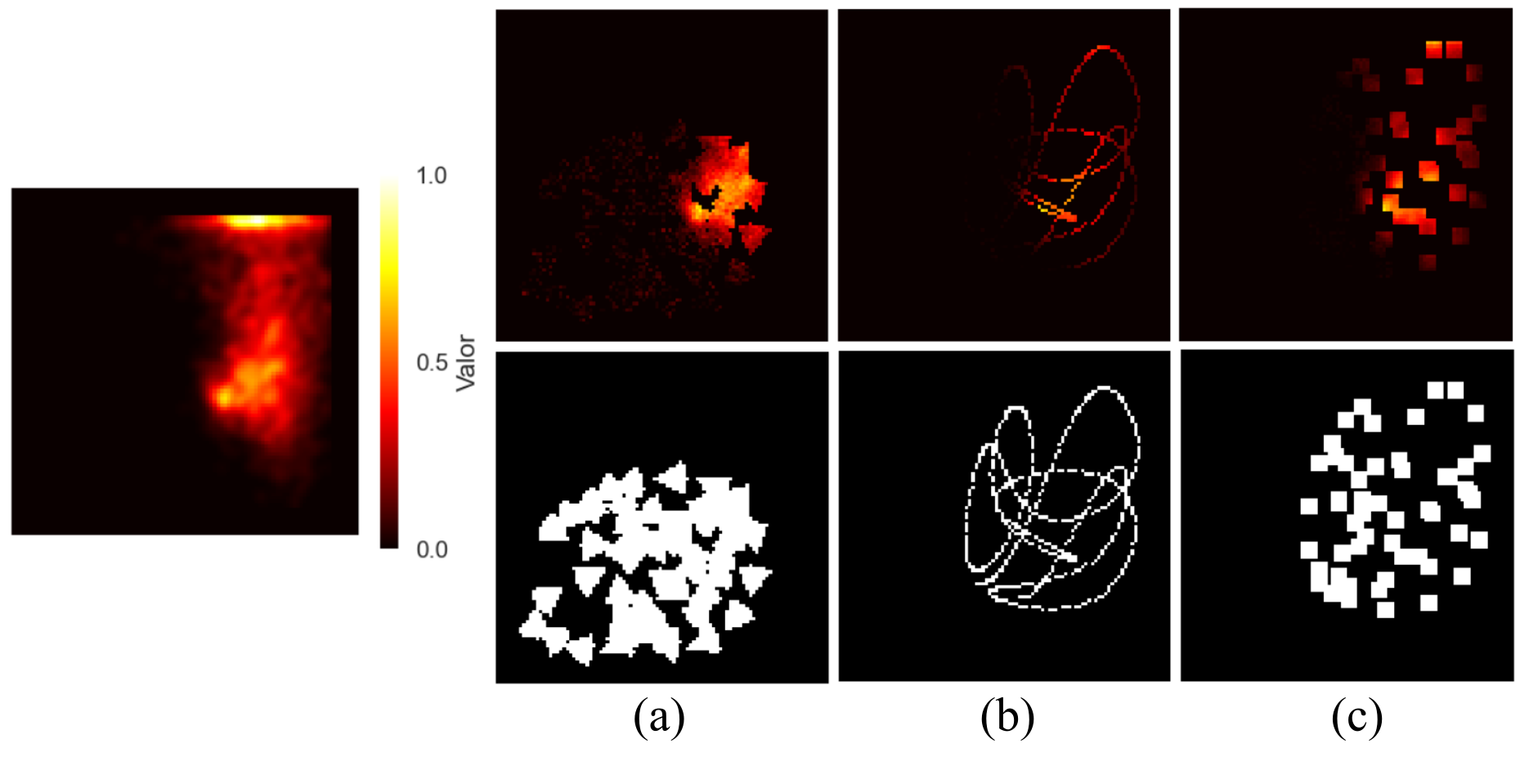}
    \caption{Example field and observation footprints for the three perceptual models for a particular scenario (left): 
    (a) Conic FOV, (b) Point-wise sensor, (c) Nadir-looking camera. 
    Top row shows the measured values, while the bottom row shows the 
    corresponding observation masks. Both matrices are fed into the U-Net.}
    \label{fig:dataset}
\end{figure}

Three perceptual models define the observation operator 
$g$ in Eq.~\eqref{eq:obs_model}, each representing a 
real sensor modality used in aquatic monitoring (see Fig. \ref{fig:dataset}).

\paragraph{Point-wise sensor.}
A contact sensor returns the field value at a single 
location: $y_i = f(\mathbf{x}_i) + \varepsilon_i$.

\paragraph{Conical field of view.}
A lateral USV camera integrates field values over a 
cone-shaped footprint:
\begin{equation}
    y_i = \frac{1}{|\mathcal{R}_i|}\sum_{\mathbf{x} 
    \in \mathcal{R}_i} f(\mathbf{x}) + \varepsilon_i,
    \label{eq:cone}
\end{equation}
where $\mathcal{R}_i$ is the set of cells within the 
conical footprint of half-angle $\phi$ and range 
$d_{\max}$ centered on the vehicle heading $\theta_i$.

\paragraph{Nadir-looking camera.}
A downward-facing UAV camera returns an $s \times s$ 
image, contributing $s^2$ spatially distributed 
observations per step:
\begin{equation}
    y_i^{(p,q)} = f\!\left(\mathbf{x}_i^{(p,q)}\right) 
    + \varepsilon_i^{(p,q)}, \qquad 
    p, q \in \{1, \ldots, s\}.
    \label{eq:nadir}
\end{equation}

\section{Results}
\label{sec:results}

\subsection{Experimental Setup}

Three synthetic datasets of 2000 oil spill simulations each 
are generated on a $100 \times 100$ grid using a 
physics-based particle simulator, differing only in their 
observation model. Base simulation parameters are fixed across 
datasets: number of particles, spill radius, wind and tidal 
speed 1.0, etc. altough every simulation generates a distinct 
field realization. The measurement noise $\varepsilon_i$ is fixed
for Nadir and Point-wise models as $\varepsilon_i \sim \mathcal{N}(0, 0.1^2)$. For
the Conical model, the noise follows a linear relationship with the pixel distance, with $\sigma_n \in [0.01, 0.08]$.
The Nadir dataset simulates a UAV with $s = 5$ and 10 m altitude, while the Conical dataset simulates a USV with $\phi = 30^\circ$ and $d_{\max} = 20$ m.
Every ground truth field is normalized to $[0,1]$ with respect to the sensor range. 
For every field, a random smooth trajectory of random length between 50 and 200 steps is generated, always within the maximum trajectory length of each vehicle.
Every trajectory is converted into a sequence of observations according to the corresponding perceptual model.
The dataset is split 80\%/20\%, yielding 400 held-out test samples per dataset. 

All deep learning models share the U-Net backbone with 
channel progression $32 \to 512$ and a bottleneck of 1024 
channels, trained for 50 epochs with batch size 16 using 
Adam. Monte Carlo Dropout uses $p = 0.2$ and $T = 30$ 
inference passes; the Ensemble comprises $M = 5$ members 
with learning rate $3 \times 10^{-4}$; EDL uses 
$\lambda = 10^{-3}$, gradient clipping at max norm 1.0, 
and learning rate $5 \times 10^{-4}$. The GP baseline 
requires no offline training and is fitted per test sample 
via L-BFGS. All experiments were conducted on a MacBook 
Pro with Apple M3 Pro and 18 GB unified RAM. Full 
implementation details and hyperparameters are available 
at \url{https://gitlab.ratatosk.cc/syanes/uncertainty-estimation-caepia-2026}.

\subsection{Evaluation Metrics}

Reconstruction accuracy is assessed using the Root Mean Squared 
Error (RMSE):

\begin{equation}
    \text{RMSE} = \sqrt{\frac{1}{|\mathcal{X}_{\text{test}}|}
    \sum_{\mathbf{x} \in \mathcal{X}_{\text{test}}} 
    \left(f(\mathbf{x}) - \hat{f}(\mathbf{x})\right)^2}.
    \label{eq:rmse}
\end{equation}

Normalized Uncertainty calibration is evaluated using the Uncertainty 
Calibration Error (UCE), which measures the consistency between 
predicted minmax-normalized uncertainty $\hat \sigma_{\text{total}}$ magnitudes and normalized empirical error $\hat e(\mathbf{x})$ variance. 
The normalized uncertainty range is discretized into $B$ equally spaced 
bins, and for each bin $U_b$ the normalized mean predicted uncertainty is 
compared against $\hat e(\mathbf{x})$:
\begin{equation}
    \text{UCE} = \sum_{b=1}^{B} \frac{|U_b|}{|\mathcal{X}_{\text{test}}|}
    \left| \frac{1}{|U_b|}\sum_{\mathbf{x} \in U_b} 
    \hat{\sigma}^2_{\text{total}}(\mathbf{x}) - 
    \frac{1}{|U_b|}\sum_{\mathbf{x} \in U_b} 
    \hat e(\mathbf{x})^2 
    \right|,
    \label{eq:uce}
\end{equation}
The normalization imposes a common scale across methods and datasets, 
enabling direct comparison of the uncertainty as a surrogate for error magnitude. 
A lower UCE indicates better calibration, with 0 representing perfect alignment between 
predicted uncertainty and empirical error.

\subsection{Experiment results}

Table~\ref{tab:rmse} reports the RMSE across the 400 test 
samples for each method and perceptual model. EDL achieves 
the best reconstruction accuracy across all three sensor 
models by a considerable margin, with a mean RMSE roughly 
half that of the Ensemble and three to four times lower 
than GP. The Deep Ensemble ranks second, while MC Dropout 
and GP trail significantly. Notably, reconstruction 
performance is largely consistent across perceptual models 
for all deep learning methods, suggesting that the U-Net 
backbone adapts well to different observation footprints. 
The GP baseline shows the highest variance, particularly 
under the Nadir model, reflecting its sensitivity to 
observation sparsity.

\begin{table}[htbp]
\centering
\caption{RMSE (mean $\pm$ std) across 400 test samples 
per dataset. Best result per column in \textbf{bold}.}
\label{tab:rmse}
\begin{tabularx}{\textwidth}{lXXX}
\toprule
\textbf{Method} & \textbf{Conic FOV} & \textbf{Nadir} 
& \textbf{Pointwise} \\
\midrule
EDL        & $\mathbf{0.0274 \pm 0.0234}$ & $\mathbf{0.0293 \pm 0.0209}$ & $\mathbf{0.0272 \pm 0.0204}$ \\
Ensemble   & $0.0506 \pm 0.0238$ & $0.0570 \pm 0.0200$ & $0.0586 \pm 0.0188$ \\
MC Dropout & $0.0878 \pm 0.0222$ & $0.0915 \pm 0.0160$ & $0.0992 \pm 0.0167$ \\
GP         & $0.1146 \pm 0.0348$ & $0.1193 \pm 0.0648$ & $0.1146 \pm 0.0321$ \\
\bottomrule
\end{tabularx}
\end{table}

Table~\ref{tab:uce} reports the UCE for each method and 
perceptual model. EDL achieves the lowest UCE across all 
sensor configurations, indicating that its predicted 
uncertainty magnitudes are most consistent with the actual 
reconstruction errors. The Ensemble ranks second with 
stable and low UCE values. MC Dropout shows moderate 
calibration, while the GP baseline presents the highest 
UCE and largest variance across all methods. This behavior 
is a direct consequence of the isotropic RBF kernel 
assumption: oil spill fields exhibit spatially 
heterogeneous Lipschitz continuity — sharp gradients 
at spill boundaries coexist with smooth interior 
regions — which a stationary, isotropic kernel cannot 
capture. As a result, the GP is forced to commit to a 
single global length-scale $\ell$ that is simultaneously 
too smooth in high-gradient regions and too rough in 
homogeneous ones. Combined with the homoscedastic noise 
term $\sigma_n^2$, which imposes a spatially uniform 
aleatoric estimate regardless of local observation 
density, the GP loses the key black-box advantage of 
principled uncertainty quantification precisely where 
it is most needed.

\begin{table}[htbp]
\centering
\caption{UCE (mean $\pm$ std) across 400 test samples 
per dataset. Best result per column in \textbf{bold}.}
\label{tab:uce}
\begin{tabularx}{\textwidth}{lXXX}
\toprule
\textbf{Method} & \textbf{Conic FOV} & \textbf{Nadir} 
& \textbf{Pointwise} \\
\midrule
EDL        & $\mathbf{0.0018 \pm 0.0031}$ & $\mathbf{0.0016 \pm 0.0025}$ & $\mathbf{0.0013 \pm 0.0022}$ \\
Ensemble   & $0.0026 \pm 0.0016$ & $0.0040 \pm 0.0016$ & $0.0040 \pm 0.0017$ \\
MC Dropout & $0.0084 \pm 0.0026$ & $0.0063 \pm 0.0018$ & $0.0110 \pm 0.0029$ \\
GP         & $0.0110 \pm 0.0158$ & $0.0241 \pm 0.2443$ & $0.0096 \pm 0.0093$ \\
\bottomrule
\end{tabularx}
\end{table}

\begin{figure}[htbp]
    \centering
    \includegraphics[width=\textwidth]{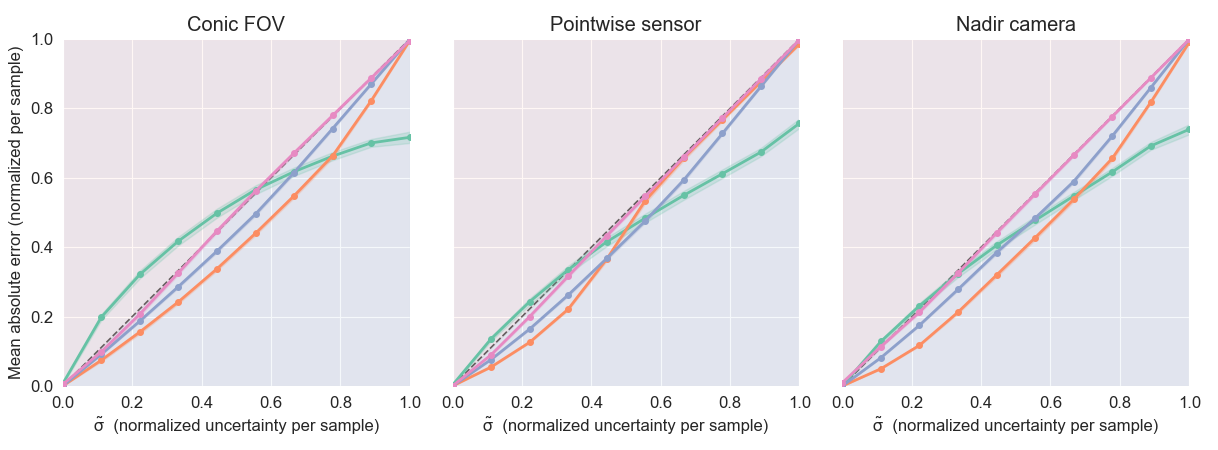}
    \caption{Calibration curves for each method and perceptual model. 
    The x-axis represents the normalized predicted uncertainty bins, 
    while the y-axis shows the corresponding normalized empirical error. 
    The closer the curve is to the diagonal, the better the calibration.}
    \label{fig:calibration}
\end{figure}

Figure~\ref{fig:calibration} shows the calibration curves 
for each method across the three perceptual models. EDL 
achieves the best calibration across all sensor models 
(deviation $\approx 0.038$--$0.040$), followed closely 
by the Ensemble (deviation $\approx 0.043$--$0.072$). 
MC Dropout presents moderate calibration errors with 
notable improvement under the Pointwise model. The GP 
baseline shows severe miscalibration across all 
configurations (deviation $> 0.52$), confirming that 
the combination of a stationary kernel and a global 
homoscedastic noise term is fundamentally misspecified 
for fields with non-homogeneous spatial regularity.

Table~\ref{tab:inference} reports mean inference time per 
sample. Beyond raw speed, inference time has direct 
implications for real-time onboard deployment. The GP 
baseline requires per-sample kernel fitting via L-BFGS, 
making its computational cost cubic in the number of 
observations: as the vehicle collects more data during 
a mission, GP inference becomes increasingly intractable, 
ranging from 81 ms under the sparse Pointwise model to 
over 2200 ms under Conic FOV, where the wider sensor 
footprint yields denser observation sets. This scaling 
behavior fundamentally limits the suitability of GPs 
for online deployment on resource-constrained platforms 
such as USVs or UAVs. Among deep learning methods, EDL is the fastest at 
$\sim$7 ms per sample, as it requires only a single 
deterministic forward pass through the network.

\begin{table}[htbp]
\centering
\caption{Mean inference time per sample (ms). 
Best result per column in \textbf{bold}.}
\label{tab:inference}
\begin{tabularx}{\textwidth}{lXXX}
\toprule
\textbf{Method} & \textbf{Conic FOV} & \textbf{Nadir} 
& \textbf{Pointwise} \\
\midrule
EDL        & $\mathbf{7.1 \pm 1.5}$  & $\mathbf{7.2 \pm 1.4}$ & $\mathbf{7.4 \pm 1.3}$ \\
Ensemble   & $20.3 \pm 2.5$          & $20.6 \pm 0.5$          & $20.9 \pm 0.6$ \\
MC Dropout & $202.4 \pm 12.4$        & $202.0 \pm 2.7$         & $207.1 \pm 8.3$ \\
GP         & $2247.1$                & $531.8$                 & $81.2$ \\
\bottomrule
\end{tabularx}
\end{table}

Taken together, the results highlight two methods as 
particularly well-suited for uncertainty-aware field 
reconstruction in autonomous vehicle deployments: 
Deep Ensembles and EDL. Both achieve strong 
reconstruction accuracy and well-calibrated uncertainty 
estimates across all perceptual models, clearly 
outperforming MC Dropout and the GP baseline.

Between the two, EDL presents a compelling additional 
advantage: it yields calibration performance comparable 
to the Ensemble at a fraction of the inference cost, 
while intrinsically modeling the predictive distribution 
as a \textit{multivariate Gaussian} through the 
Normal-Inverse-Gamma formulation. This means that 
epistemic and aleatoric uncertainty are not post-hoc 
approximations derived from sampling, but closed-form 
quantities that emerge directly from the evidential 
prior. This makes EDL particularly attractive for 
integration into probabilistic IPP frameworks where 
the uncertainty map must be queried repeatedly and 
efficiently during online mission execution.

%


\section{Conclusion}
\label{sec:conclusion}

This paper has presented a comparative evaluation of four 
uncertainty estimation methods for scalar field reconstruction 
under heterogeneous observation models, motivated by the 
requirements of uncertainty-aware Informative Path Planning 
with autonomous surface and aerial vehicles.

Evidential Deep Learning and Deep Ensembles consistently 
outperform Monte Carlo Dropout and the Gaussian Process 
baseline across all perceptual models. Evidential Deep 
Learning achieves the lowest RMSE and UCE with a single 
deterministic forward pass, while Deep Ensembles offer 
competitive calibration at a modest increase in inference 
cost. Crucially, both methods produce uncertainty estimates 
that are well aligned with the actual reconstruction error, 
making predicted uncertainty a reliable surrogate for the 
true error surface. This calibration property is directly 
exploitable in path planning strategies that optimize 
reconstruction quality without access to ground truth, 
and opens the door to online model selection criteria 
grounded in uncertainty rather than held-out performance.

The Gaussian Process baseline, despite its principled 
formulation, suffers from severe miscalibration across 
all configurations due to its stationary isotropic kernel, 
which cannot adapt to fields with spatially heterogeneous 
regularity. Its cubic scaling with observation count 
further limits its suitability for real-time onboard 
deployment.

Future work will integrate these methods into a closed-loop 
Informative Path Planning system to evaluate whether 
well-calibrated epistemic uncertainty translates into 
improved path efficiency under budget-constrained missions.

\begin{credits}
    \subsubsection{\ackname} Proyecto PID2024-158365OB-C21 financiado por \\MICIU/AEI/10.13039/501100011033 y por FEDER, UE.

    \subsubsection{\discintname}
    The authors have no competing interests to declare that are
    relevant to the content of this article.
\end{credits}
%
%
%
\bibliographystyle{splncs04}
\bibliography{bibliography}

@article{zakaria2021uav,
  author    = {Zakaria, Noor Atikah and others},
  title     = {{UAV}-based remote sensing for the petroleum industry and
               environmental monitoring: State-of-the-art and perspectives},
  journal   = {Journal of Petroleum Science and Engineering},
  volume    = {208},
  pages     = {109633},
  year      = {2022},
  publisher = {Elsevier}
}

@article{chen2019robotic,
  author    = {Chen, Weizhe and Khardon, Roni and Liu, Lantao},
  title     = {Robotic Active Information Gathering for Spatial Field
               Reconstruction with Rapidly-Exploring Random Trees and Online
               Learning of {G}aussian Processes},
  journal   = {Sensors},
  volume    = {19},
  number    = {5},
  pages     = {1016},
  year      = {2019},
  publisher = {MDPI}
}

@article{mansfield2024survey,
  author    = {Mansfield, Samuel and Montazeri, Allahyar},
  title     = {A survey on autonomous environmental monitoring approaches:
               towards unifying active sensing and reinforcement learning},
  journal   = {Frontiers in Robotics and {AI}},
  volume    = {11},
  pages     = {1336612},
  year      = {2024},
  publisher = {Frontiers}
}

@inproceedings{gal2016dropout,
  author    = {Gal, Yarin and Ghahramani, Zoubin},
  title     = {Dropout as a {B}ayesian Approximation: Representing Model
               Uncertainty in Deep Learning},
  booktitle = {Proceedings of the 33rd International Conference on Machine
               Learning ({ICML})},
  pages     = {1050--1059},
  year      = {2016}
}

@inproceedings{laksh2017deep,
  author    = {Lakshminarayanan, Balaji and Pritzel, Alexander and
               Blundell, Charles},
  title     = {Simple and Scalable Predictive Uncertainty Estimation using
               Deep Ensembles},
  booktitle = {Advances in Neural Information Processing Systems~30
               ({NeurIPS})},
  pages     = {6405--6416},
  year      = {2017}
}

@inproceedings{amini2020deep,
  author    = {Amini, Alexander and Schwarting, Wilko and Soleimany, Ava
               and Rus, Daniela},
  title     = {Deep Evidential Regression},
  booktitle = {Advances in Neural Information Processing Systems~33
               ({NeurIPS})},
  year      = {2020}
}

@inproceedings{nix1994estimating,
  author    = {Nix, David A. and Weigend, Andreas S.},
  title     = {Estimating the mean and variance of the target
               probability distribution},
  booktitle = {Proceedings of the 1994 {IEEE} International Conference
               on Neural Networks ({ICNN})},
  volume    = {1},
  pages     = {55--60},
  year      = {1994},
  publisher = {IEEE}
}

@article{samaniego2021bayesian,
  title={A bayesian optimization approach for water resources monitoring through an autonomous surface vehicle: The ypacarai lake case study},
  author={Samaniego, Federico Peralta and Reina, Daniel Guti{\'e}rrez and Mar{\'\i}n, Sergio L Toral and Arzamendia, Mario and Gregor, Derlis O},
  journal={IEEE Access},
  volume={9},
  pages={9163--9179},
  year={2021},
  publisher={IEEE}
}

@article{syanes2023censored,
title = {Censored deep reinforcement patrolling with information criterion for monitoring large water resources using Autonomous Surface Vehicles},
journal = {Applied Soft Computing},
volume = {132},
pages = {109874},
year = {2023},
issn = {1568-4946},
doi = {10.1016/j.asoc.2022.109874},
author = {Samuel {Yanes Luis} and Daniel Gutiérrez-Reina and Sergio {Toral Marín}}
}

@article{syanes2024deep,
author = {Yanes Luis, Samuel and Shutin, Dmitriy and Marchal Gómez, Juan and Gutiérrez Reina, Daniel and Toral Marín, Sergio},
title = {Deep Reinforcement Multiagent Learning Framework for Information Gathering with Local Gaussian Processes for Water Monitoring},
journal = {Advanced Intelligent Systems},
volume = {6},
number = {8},
pages = {2300850},
doi = {10.1002/aisy.202300850},
year = {2024}
}

@Article{vae,
AUTHOR = {Casado-Pérez, Alejandro and Yanes, Samuel and Toral, Sergio L. and Perales-Esteve, Manuel and Gutiérrez-Reina, Daniel},
TITLE = {Variational Autoencoder for the Prediction of Oil Contamination Temporal Evolution in Water Environments},
JOURNAL = {Sensors},
VOLUME = {25},
YEAR = {2025},
NUMBER = {6},
ARTICLE-NUMBER = {1654},
PubMedID = {40292734},
ISSN = {1424-8220},
DOI = {10.3390/s25061654}
}

@misc{UNET,
      title={U-Net: Convolutional Networks for Biomedical Image Segmentation}, 
      author={Olaf Ronneberger and Philipp Fischer and Thomas Brox},
      year={2015},
      eprint={1505.04597},
      archivePrefix={arXiv},
      primaryClass={cs.CV},
      url={https://arxiv.org/abs/1505.04597}, 
}

@book{rasmussen2006gaussian,
  author    = {Rasmussen, Carl Edward and Williams, Christopher K. I.},
  title     = {Gaussian Processes for Machine Learning},
  publisher = {The MIT Press},
  year      = {2006},
  doi       = {10.7551/mitpress/3206.001.0001},
  url       = {https://doi.org/10.7551/mitpress/3206.001.0001},
  isbn      = {026218253X},
  series    = {Adaptive Computation and Machine Learning}
}

\end{document}